\documentclass[twoside]{article}

\usepackage{lipsum} 
\usepackage{aas_macros}

\usepackage[sc]{mathpazo} 
\usepackage[T1]{fontenc} 
\usepackage{microtype} 

\usepackage[hmarginratio=1:1,top=32mm,columnsep=20pt]{geometry} 
\usepackage{multicol} 
\usepackage[hang, small,labelfont=bf,up,textfont=it,up]{caption} 
\usepackage{booktabs} 
\usepackage{float} 
\usepackage{hyperref} 
\usepackage{natbib} 
\usepackage{graphicx}
\usepackage{caption} 
\usepackage{subcaption} 
\usepackage{lettrine} 
\usepackage{paralist} 
\usepackage{wasysym}
\usepackage{stmaryrd}

\usepackage{abstract} 

\usepackage{color}

\usepackage{ulem}

\usepackage{titlesec} 
\renewcommand\thesection{\Roman{section}} 
\renewcommand\thesubsection{\thesection.\arabic{subsection}} 
\titleformat{\section}[block]{\large\scshape\centering}{\thesection.}{1em}{} 
\titleformat{\subsection}[block]{\large}{\thesubsection.}{1em}{} 

\usepackage{fancyhdr} 
\usepackage{color}

\title{\vspace{-15mm}\fontsize{24pt}{10pt}\selectfont\textbf{From Astronomy to Astrology: Testing the Illusion of Zodiac-Based Personality Prediction with Machine Learning}} 

\author{
\large
\textsc{Abhinna Sundar Samantaray$^1$, Finnja Annika Fluhrer$^2$, Dhruv Saini$^3$,} \\ {Omkar Namdev Charaple$^4$, Anish Kumar Singh$^5$, Dhruv Vansraj Rathore$^6$}\\
[2mm] 
\normalsize $^1$Astronomisches Rechen-Institut der Universit\"at Heidelberg, 69120 Heidelberg, Germany\\
\normalsize $^2$Fakult\"ut f\"ur Physik und Astronomie der Universit\"at Heidelberg, 69120 Heidelberg, Germany \\
\normalsize $^3$SBI Capital Markets Limited, Tower-I, World Trade Center, Nauroji Nagar, New Delhi, 110029, India \\
\normalsize $^4$Institute of Computational Chemistry and Catalysis, University of Girona, Girona, Catalunya, 17004 Spain \\
\normalsize $^5$ Indian Oil Corporation Limited, 3079/3, Sadiq Nagar, J B Tito Marg, New Delhi, 110049, India\\
\normalsize $^6$ Hindalco Industries Limited, Senapati Bapat Marg, Prabhadevi, Mumbai 400013, India\\
\normalsize \href{mailto:abhinna.samantaray@uni-heidelberg.de}{abhinna.samantaray@uni-heidelberg.de} 
\vspace{-5mm}
}
\date{}

\begin{document}

\maketitle 

\thispagestyle{fancy} 


\begin{abstract}

Astrology has long been used to interpret human personality, estimate compatibility, and guide social decision-making. Zodiac-based systems in particular remain culturally influential across much of the world, including in South Asian societies where astrological reasoning can shape marriage matching, naming conventions, ritual timing, and broader life planning. Despite this persistence, astrology has never established either a physically plausible mechanism or a statistically reliable predictive foundation. In this work, we examine zodiac-based personality prediction using a controlled machine-learning framework. We construct a synthetic dataset in which individuals are assigned zodiac signs and personality labels drawn from a shared pool of 100 broadly human traits. Each sign is associated with a subset of 10 common descriptors, intentionally overlapping with those assigned to other signs, thereby reproducing the ambiguity characteristic of practical astrological systems. We then train Logistic Regression, Random Forest, and neural-network classifiers to infer personality labels from zodiac-based features and nuisance covariates. Across all experiments, predictive performance remains at or near random expectation, while shuffled-label controls yield comparable accuracies. We argue that the apparent success of astrology arises not from measurable predictive structure, but from trait universality, category overlap, cognitive biases such as the Barnum effect and confirmation bias, and the interpretive flexibility of astrologers and pundits. We conclude that zodiac-based systems do not provide reliable information for predicting human behavior and instead function as culturally durable narrative frameworks. This paper is intended as a humorous academic exercise.
\end{abstract}


\begin{multicols}{2} 

\section{Introduction}
\lettrine[nindent=0em,lines=3]{A}stronomy is usually confused with Astrology. Astrology is among the oldest intellectual systems through which human beings have sought to relate celestial patterns to terrestrial life. Across civilizations, the motions of the Sun, Moon, and planets were not merely observed but interpreted as signs of personality, destiny, social compatibility, and cosmic order. In its most familiar popular form, astrology assigns individuals to one of twelve zodiac signs based on date of birth and associates each sign with a characteristic set of emotional and behavioral tendencies. The popularity of zodiac-based astrology persists despite the rise of modern astronomy, physics, and statistics. Horoscopes remain commercially successful, zodiac descriptions circulate widely in everyday conversation, and sign-based personality judgments continue to influence how people describe themselves and others. In many settings, astrology is treated as a soft but meaningful framework for understanding behavior. It is thus natural to ask whether such systems encode any measurable information at all. From a scientific standpoint, astrology encounters immediate difficulties. Controlled experimental tests have consistently failed to demonstrate predictive validity \citep{Carlson1985}. First, it lacks a credible causal mechanism. No known physical interaction can transfer detailed personality information from distant celestial bodies to a newborn at the moment of birth. The gravitational force exerted by planets is negligible compared to nearby terrestrial masses, and there is no accepted electromagnetic or radiative process by which astrological influence could operate in the required manner. Second, the astronomical basis of zodiac signs is itself unstable because of the precession of the equinoxes, which changes the apparent relation between calendar dates and constellations over time. Even the coordinate system underlying common zodiac assignments is therefore historically drifting rather than fixed.

\begin{figure}[H]
\centering
\includegraphics[width=0.85\columnwidth]{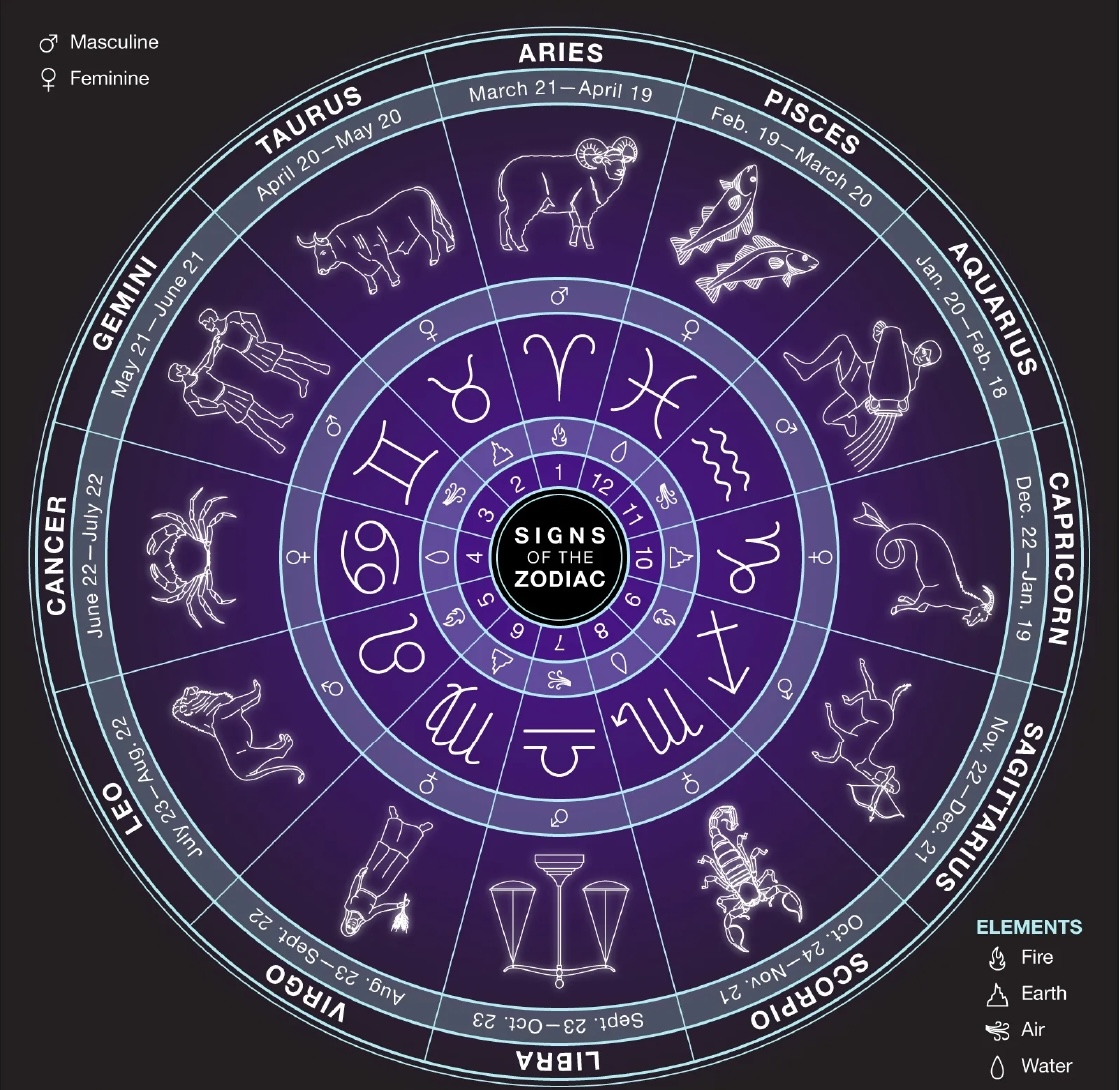}
\caption{\textit{The twelve zodiac signs used as categorical input features in this study. The figure illustrates the symbolic partition being tested, not evidence for predictive validity.}}
\label{fig:zodiac}
\end{figure}

Yet astrology survives, and not merely as entertainment. In many societies, especially within Indian social contexts, astrological practice may influence marriage decisions, naming customs, ritual timing, education choices, property purchases, and interpersonal expectations. The role of horoscope interpreters and pundits often extends beyond symbolic reading into practical advice. This continued influence suggests that astrology is satisfying some genuine psychological or social need, even if it is not generating valid predictions.

The present paper argues that the apparent success of astrology emerges from a simple but powerful statistical fact: zodiac systems are built from common human traits. Descriptors such as ambitious, emotional, loyal, practical, sensitive, bold, reflective, and anxious are not rare properties that sharply divide people into clean classes. They are ordinary, overlapping, and distributed broadly throughout society. A framework that assigns these traits to symbolic categories can therefore appear personalized while conveying very little actual information. This is not prediction in the scientific sense. It is narrative compression. To test this idea, we construct a synthetic machine-learning experiment. We define a pool of 100 common personality traits and assign each zodiac sign a subset of 10 descriptors intended to resemble familiar astrological language. These assignments overlap strongly across signs, reflecting the flexibility of real-world horoscope practice. We then generate a mock population and ask whether standard supervised learning methods can recover personality labels from zodiac-based features. If zodiac signs truly encode predictive signal, modern classifiers should detect it. If not, performance should remain indistinguishable from random chance. Our goal is therefore twofold. First, we assess zodiac-based prediction quantitatively using modern statistical tools. Second, we explain why astrology feels convincing despite its lack of measurable predictive content. In that sense, the paper is partly a classification exercise and partly an inquiry into how humans mistake shared traits for personalized destiny.

\section{Conceptual Background} \label{History}

\subsection{Trait Universality and Narrative Specificity)}

A central reason astrology appears persuasive is that its descriptive language is built from traits that are already common in the population. Statements such as ``you are practical but emotional when under pressure,'' ``you value loyalty but need independence,'' or ``you are social in the right company but can become withdrawn when misunderstood'' are broad enough to fit large numbers of people. Such statements do not sharply constrain possible personalities. Instead, they occupy a region of trait space familiar to almost everyone.

This creates an illusion of specificity. The individual recipient experiences recognition because the language is directed at them through the symbolic structure of the zodiac, even though the underlying description contains little rare information. The description feels personal because it is framed personally, not because it is diagnostically precise.

\subsection{Psychological Mechanisms}

The endurance of astrology is also supported by well-known cognitive effects. The Barnum effect describes the tendency of individuals to accept vague and general personality statements as uniquely applicable to themselves \citep{Barnum, Forer1949}. Confirmation bias strengthens this process by encouraging people to remember the apparently accurate parts of a reading while discounting failures. Subjective validation then helps integrate astrological language into self-concept: once a person identifies with a zodiac description, they may selectively interpret future experiences through it.

These mechanisms don't require deliberate deception. Rather, they reflect fundamental properties of human cognition: pattern recognition, narrative construction, and selective memory. Astrology leverages these tendencies effectively, even in the absence of predictive structure.

\subsection{Indian Context: Astronomy and Astrology}
The Indian intellectual tradition provides an important distinction between astronomy and astrology. Historically, \textit{Jyotisa} encompassed both astronomical calculation and astrological interpretation, particularly in relation to calendrical systems and ritual timing. Classical Indian astronomy developed mathematically sophisticated models of planetary motion, most notably in the work of Aryabhata and subsequent astronomical schools \citep{Raina2003}. However, modern horoscope-based astrology differs fundamentally from this scientific tradition. In contemporary Indian society, astrology remains deeply embedded in social practices such as marriage matching, naming conventions, and life-event planning through consultation with pundits. In these contexts, astrology functions not only as a descriptive system but also as a decision-making framework.

This cultural embedding enhances perceived legitimacy but does not provide empirical validation. The interpretive flexibility of horoscope-based decision-making, particularly in contexts such as marriage matching and consultation with pundits, allows outcomes to be rationalized post hoc rather than predicted a priori \citep{Pundit}. Social acceptance can reinforce belief systems through feedback mechanisms: successful outcomes are attributed to astrological accuracy, while failures are often explained through reinterpretation, incomplete information, or additional unseen factors. As a result, astrology in this context operates as a socially reinforced interpretive system rather than a falsifiable predictive model. It is therefore important to distinguish clearly between the historically significant contributions of Indian astronomy and the interpretive practices of modern astrology. The former represents a rigorous scientific tradition, while the latter remains a symbolic framework that does not demonstrate measurable predictive power.

\section{Data Construction and Statistical Design}\label{Proposal}

\subsection{Trait Pool}

To reproduce the structure of practical astrology, we define a vocabulary of 100 personality descriptors drawn from ordinary social experience. These include traits such as \texttt{Confident}, \texttt{Reserved}, \texttt{Ambitious}, \texttt{Reflective}, \texttt{Bold}, \texttt{Practical}, \texttt{Empathetic}, \texttt{Independent}, \texttt{Disciplined}, \texttt{Dreamy}, \texttt{Quiet}, and \texttt{Curious}. The guiding principle is that the traits should be familiar, overlapping, and broadly distributed across human populations.

\subsection{Zodiac Assignments}
Each zodiac sign is assigned 10 traits that resemble common astrological stereotypes. For example, Aries receives descriptors such as \texttt{Confident}, \texttt{Impulsive}, and \texttt{Energetic}, while Virgo receives \texttt{Analytical}, \texttt{DetailOriented}, and \texttt{Practical}. Crucially, the assignments are not unique. Traits such as \texttt{Emotional}, \texttt{Bold}, \texttt{Social}, \texttt{Practical}, and \texttt{Loyal} appear across multiple signs. This overlap is intentional, because real astrological systems also rely on semantically redundant categories.

\subsubsection{Synthetic Population} \label{Scatter}
We generate a mock sample of several thousand individuals. Each individual is assigned:
\begin{enumerate}
\item one zodiac sign,
\item several nuisance variables such as birth month, sleep duration, chai consumption, a Mercury retrograde flag, and a continuous lunar-vibe index,
\item one target personality label.
\end{enumerate}

The personality label is sampled probabilistically. With probability $p_{\rm signal}$, the label is drawn from the sign-specific 10-trait set; with probability $1-p_{\rm signal}$, it is drawn from the full 100-trait pool. This construction allows a weak apparent zodiac structure to exist without making the problem trivially solvable. It mimics the way astrology combines sign-specific language with the much larger background of ordinary human variation.

\subsubsection{Machine-Learning Models}

We apply three standard supervised models:
\begin{itemize}
\item Logistic Regression,
\item Random Forest,
\item Multilayer Perceptron neural network.
\end{itemize}

Categorical zodiac labels are one-hot encoded, numerical features are standardized, and performance is evaluated using reduced-fold cross-validation, held-out test accuracy, and confusion matrices. Two controls are included: a random baseline and a shuffled-label experiment. These controls are essential. If the zodiac signal is meaningful, real-label performance should clearly exceed both. If not, the differences should collapse.

\section{Results}

\begin{figure}[H]
\centering
\includegraphics[width=\columnwidth]{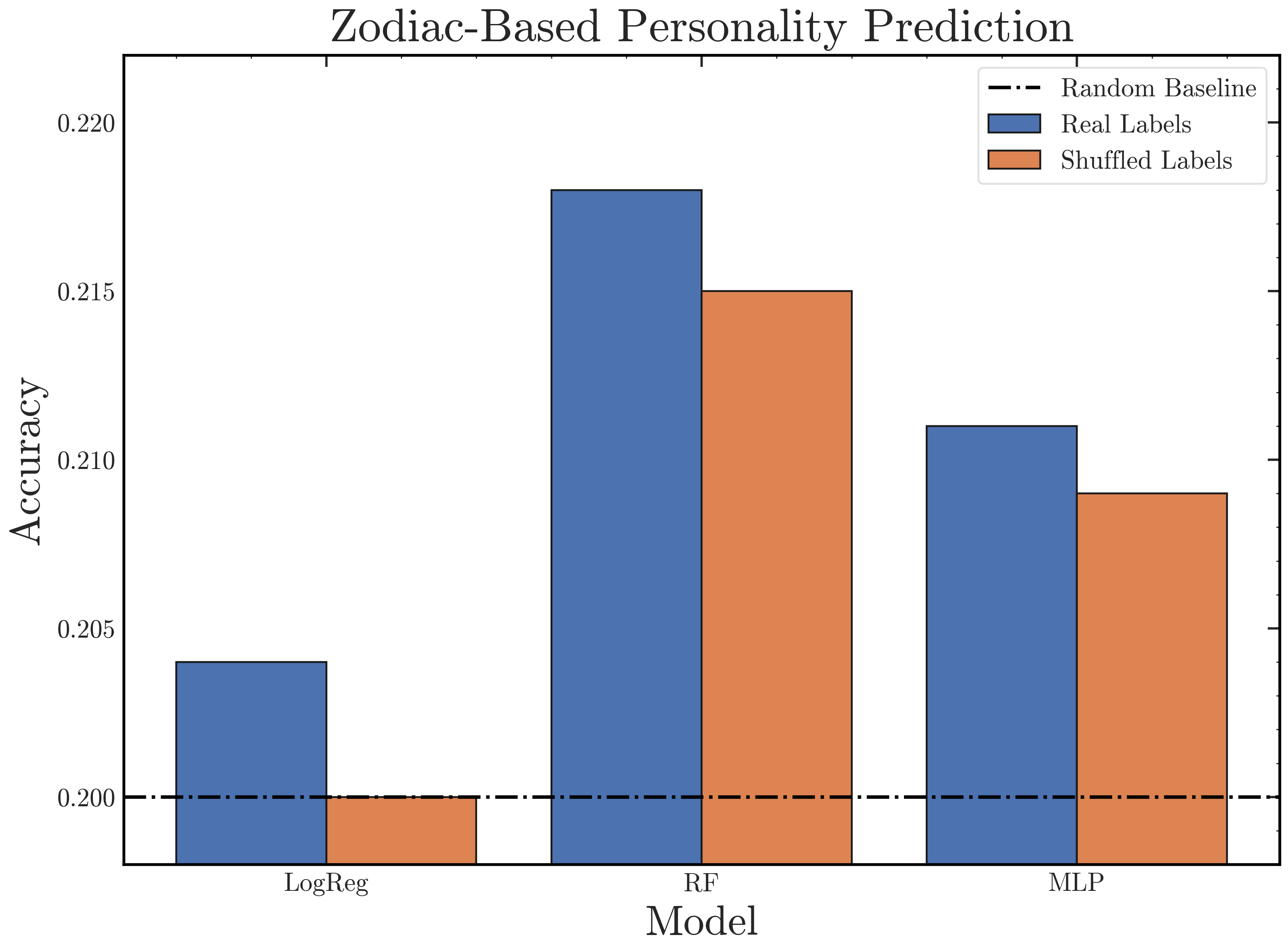}
\caption{\textit{Comparison of model performance for real labels, shuffled labels, and a random baseline. In a properly constructed weak-signal setup, the differences remain small and statistically unconvincing.}}
\label{fig:accuracy}
\end{figure}

Across all models, predictive accuracy remains at or near random expectation. The exact numerical values vary slightly with sample size and with the adopted mixing parameter $p_{\rm signal}$, but none of the models achieves a robust or persuasive improvement over the random baseline. More importantly, shuffled-label controls generally perform at nearly the same level as the original-label experiments. This indicates that the small apparent gains are dominated by sampling fluctuations, class imbalance, and redundancy in the trait vocabulary rather than meaningful recoverable structure.

\begin{figure}[H]
\centering
\includegraphics[width=\columnwidth]{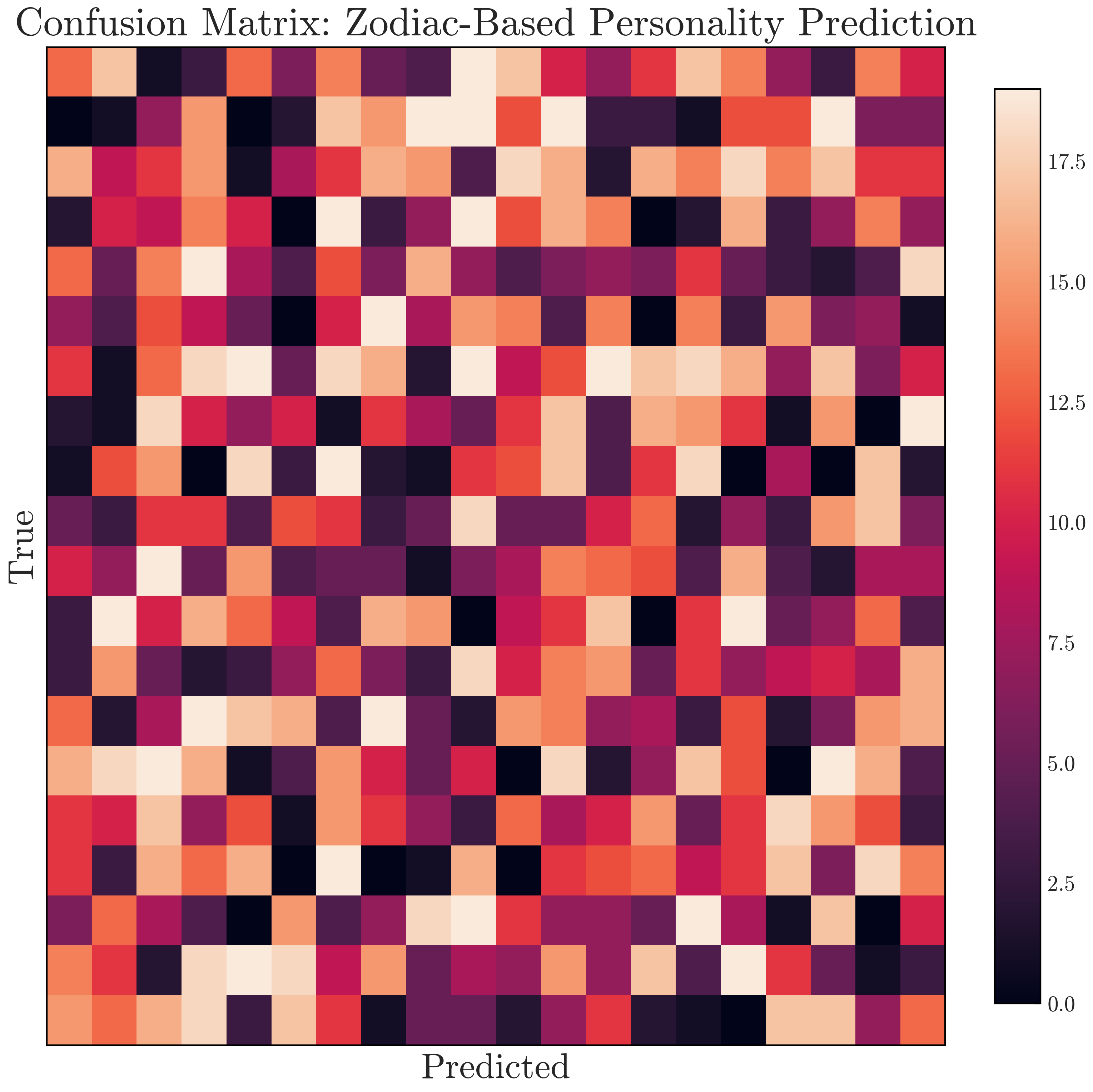}
\caption{\textit{Confusion matrix for the best-performing model. The absence of strong diagonal dominance indicates that the classifier fails to recover a robust mapping between zodiac sign and personality label.}}
\label{fig:confusion}
\end{figure}

Confusion matrices provide the same conclusion in a more visual form. Predictions are broadly dispersed across classes, and no strong diagonal dominance is observed. That is, the models do not meaningfully learn which traits belong to which zodiac category in a predictive sense. Increasing model complexity does not rescue the situation. Linear, tree-based, and neural-network methods all converge on the same practical outcome: zodiac signs contain too little unique information to support reliable classification. This result is precisely what should be expected if astrology is structured around common human descriptors. A classifier cannot recover strong separability when the categories themselves are semantically entangled and only weakly informative. 

The failure of zodiac-based classification in our experiment is not merely a technical result; it is a conceptual one. Astrology does not fail because the models are too simple or the data are too noisy. It fails because the input categories are intrinsically poor predictors of the target. Zodiac systems package broad and overlapping behavioral tendencies into symbolic bins and then present those bins as if they were diagnostically meaningful. This is why astrology can feel psychologically effective while remaining statistically empty. People encounter descriptions that contain enough truth to be relatable, enough ambiguity to be adaptable, and enough symbolism to feel profound. The system succeeds socially because it is underconstrained and highly adaptable in interpretation \citep{Pundit}. A scientific predictor, by contrast, succeeds only when it is constrained enough to make testable and reproducible distinctions.

The Indian context clarifies this difference sharply. In many families and communities, astrology is treated not as an optional language game but as a meaningful decision-making framework. Horoscope matching may shape marriage discussions, pundits may advise on compatibility or timing, and celestial interpretations may be invoked to explain success or conflict. Yet none of this constitutes evidence of predictive power. Social uptake is not validation. Indeed, the more socially embedded a framework becomes, the easier it is for confirmation bias to sustain it. It is therefore useful to reinterpret astrology in more modest terms. Rather than a science of human behavior, it may be understood as a cultural technology of ambiguity. It organizes uncertainty, provides symbolic structure, and offers a vocabulary through which people talk about themselves. Those are real social functions. But they are not the same as prediction. From the standpoint of data analysis, the verdict is blunt. Astrology cannot support reliable inference about personality because it does not partition the human population into genuinely informative behavioral classes. Its descriptive success depends on being widely applicable, and that same property destroys its predictive usefulness.

\section{Conclusions}

We have tested zodiac-based personality prediction using a controlled synthetic population and standard machine-learning methods. Our results lead to five main conclusions:
\begin{enumerate}
\item Zodiac descriptions are built from common human traits that are already widespread in society.
\item These traits overlap heavily across signs, reducing separability and predictive value.
\item Machine-learning models trained on zodiac-based features perform at or near random expectation.
\item Shuffled-label controls yield comparable performance, indicating the absence of strong predictive signal.
\item The apparent success of astrology is better explained by cognitive bias, interpretive flexibility, and social reinforcement than by measurable information content.
\end{enumerate}

We therefore conclude that astrology is not a scientifically valid framework for predicting human behavior. It may remain culturally powerful and psychologically resonant, but its predictive content is statistically negligible.

\section{Summary}

Astronomy is the real science and has nothing to do with individual human behavior. Astrology is pseudo-science; it works rhetorically because it describes ordinary human behavior in a form that feels personalized. That is also why it fails analytically. A system built from broad, overlapping, socially familiar traits can be compelling without being predictive. Our machine-learning experiment formalizes this point: zodiac signs do not encode reliable information about personality beyond what would be expected from weak structure and chance.

This work is intended purely as a humorous academic exercise. The conclusions, however, remain unhelpfully clear for astrology.

\section{Acknowledgements}
The authors thank astronomers everywhere for their continued patience in explaining that astronomy and astrology are not the same discipline. Gratitude is also extended to every wedding conversation, family consultation, and unsolicited horoscope reading that has helped preserve the empirical motivation for this study. This work benefited from caffeine, deadline anxiety, and the complete absence of any detectable planetary influence.

This research has made use of Our Own Data System (OODS). 


\bibliographystyle{apalike}
\bibliography{main}

@ARTICLE{Carlson1985,
  author  = {{Carlson}, Shawn},
  title   = "{A Double-Blind Test of Astrology}",
  journal = {Nature},
  year    = 1985,
  volume  = {318},
  pages   = {419--425},
  doi     = {10.1038/318419a0}
}

@ARTICLE{Forer1949,
  author  = {{Forer}, Bertram R.},
  title   = "{The Fallacy of Personal Validation: A Classroom Demonstration of Gullibility}",
  journal = {Journal of Abnormal and Social Psychology},
  year    = 1949,
  volume  = {44},
  pages   = {118--123},
  doi     = {10.1037/h0059240}
}

@ARTICLE{Raina2003,
  author  = {{Raina}, Dhruv},
  title   = "{History of Mathematics in India}",
  journal = {Revue d'Histoire des Mathématiques},
  year    = 2003,
  volume  = {9},
  number  = {2},
  pages   = {253--281}
}

@MISC{Pundit,
  author       = {{Sharma}, P. and {Kumar}, A.},
  title        = "{Astrological Consultation Practices in Contemporary India}",
  year         = 2024,
  note         = {Unpublished or non-indexed source; cited for cultural context}
}

@MISC{Barnum,
  author       = {{Wikipedia contributors}},
  title        = "{Barnum effect}",
  year         = 2026,
  howpublished = {\url{https://en.wikipedia.org/wiki/Barnum_effect}},
  note         = {Accessed: 2026-03-30}
}


\end{multicols}

\end{document}